# Patient-Centered, Graph-Augmented Artificial Intelligence-Enabled Passive Surveillance for Early Stroke Risk Detection in High-Risk Individuals

Jiyeong Kim, PhD, MPH[1], Stephen P. Ma, MD, PhD[2], Nirali Vora, MD[3], Nicholas W. Larsen, MD[3], Julia Adler-Milstein, PhD[4], Jonathan H. Chen[2,5], Selen Bozkurt, PhD, MS[6], Abeed Sarker, PhD[6], Juhee Cho, PhD[7], Jindeok Joo, MD, PhD[8,9], Natali Pageler, MD, MPH[10], Fatima Rodriguez, MD, MPH[11], Christopher Sharp, MD[2], Eleni Linos, MD, DrPH[1]

[1] Stanford Center for Digital Health, School of Medicine, Stanford University, Stanford, CA, USA
[2] Hospital Medicine, Department of Medicine, Stanford School of Medicine, Stanford, CA, USA
[3] Department of Neurology and Neurological Sciences, School of Medicine, Stanford University, Stanford, CA, USA
[4] Division of Clinical Informatics and Digital Transformation, Department of Medicine, University of California, San Francisco, CA, USA
[5] Division of Computational Medicine, School of Medicine, Stanford University, Stanford, CA, USA
[6] Department of Biomedical Informatics, School of Medicine, Emory University, Atlanta, GA, USA
[7] Department of Clinical Research Design and Evaluation, Sungkyunkwan University School of Medicine, Seoul, South Korea
[8] Department of Neurosurgery, School of Medicine, Stanford University, Stanford, CA, USA
[9] Department of Neurosurgery, Jeju National University Hospital, Jeju, South Korea
[10] Division of Clinical Informatics, School of Medicine, Stanford University, Stanford, CA, USA
[11] Division of Cardiovascular Medicine and Cardiovascular Institute, Department of Medicine, Stanford University, Stanford, CA, USA

**Abstract**

Stroke affected millions annually, yet poor symptom recognition often delayed care-seeking. To address risk recognition gap, we developed a passive surveillance system for early stroke risk detection using patient-reported symptoms among individuals with diabetes. Constructing a symptom taxonomy grounded in patients' own language and a dual machine learning pipeline (heterogeneous GNN and EN/LASSO), we identified symptom patterns associated with subsequent stroke. We translated findings into a hybrid risk screening system integrating symptom relevance and temporal proximity, evaluated across 3-90 day windows through EHR-based simulations. Under conservative thresholds, intentionally designed to minimize false alerts, the screening system achieved high specificity (1.00) and prevalence-adjusted positive predictive value (1.00), with good sensitivity (0.72), an expected trade-off prioritizing precision, that was highest in 90-day window. Patient-reported language alone supported high-precision, low-burden early stroke risk detection, that could offer a valuable time window for clinical evaluation and intervention for high-risk individuals.

[146/150 words]

**Introduction**

Stroke affected over 9.04 million people in 2020, with healthcare spending exceeding.[1,2] As the fifth leading U.S. cause of death, stroke remains a major driver of long-term physical, cognitive, and emotional disability.[3] Rapid recognition within hours facilitates emergency intervention, that can dramatically improve survival and functional recovery by enabling thrombolytic or endovascular therapies.[4] Despite advances in acute stroke care, prehospital delays continue to limit these benefits, with nearly half of patients arriving too late for optimal treatment.

Patient-level barriers are a major contributor to these delays, even among the highest-risk patients although broader ecological and socially vulnerability played a role in prehospital delay.[5] People with diabetes face a 2-4-fold higher risk of stroke, yet may misattribute stroke-like symptoms, such as assuming dizziness or numbness is due to fatigue, medication, or neuropathy. They may not recognize symptoms represent an emergency and may contact family members or outpatient clinicians rather than paramedics.[4] Health literacy gaps also play a central role in marginalized populations, three in five individuals with diabetes cannot name even two warning signs of stroke.[6] Even when patients are aware of stroke signs, uncertainty and denial can hinder action, particularly when symptoms are transient, mild, or fall outside the FAST (Face-Arm-Speech-Time) framework. Collectively, these patient-level barriers highlight the need for new approaches to enhance stroke symptom recognition and empower timely responses.

Patient portals enable secure, asynchronous patient-reported symptom sharing and often serve as the first point of contact for high risk patients when new concerns arise. Patients with diabetes are among the most active portal users.[7] A previous study, using natural language processing (NLP) to analyze secure messages from patients with diabetes, found that patients frequently described a variety of symptoms in their own words (e.g., vision changes, bone pain, and sleep problems) with timestamps.[8] These messages could offer a valuable, real-world signal that can be leveraged for diverse symptom detection. Despite the novel value of secure patient messages, symptoms reported in patient messages have been underutilized, potentially due to the challenges in structuring disparate free-text messages into clinically actionable information.

Graph neural networks (GNNs) can capture complex dependencies in a tree-like data structure.[9] Due to its known powerful ability in classification and prediction from large heterogeneous data, highly subjective to individuals, such as patient-reported symptoms, it has been used for disease detection and prognostic prediction.[10] Moreover, with recent advancements in natural language understanding, large language models (LLMs) have shown promising performance in detecting symptoms from patient messages.[11] Although the languages are eclectic and the contextual information could be limited in patient messages, LLMs were able to distinguish at-risk individuals with disease.[12]

Therefore, we aimed to address a critical gap in early risk recognition by leveraging the novel value of patient-reported signs from secure messages and harnessing the capacity of machine learning (ML) and LLMs to develop a patient-centered passive surveillance for early stroke risk detection. We hypothesized that our approach could detect stroke at-risk individuals early by screening clinically meaningful stroke-related symptoms from secure patient messages. Hence, first, we created a comprehensive and hierarchical symptom map grounded in patients' own language and applied GNNs to identify symptom patterns statistically and temporally associated with subsequent stroke. We, then, evaluated clinical utility of this approach by simulating LLM-guided symptom screening within the electronic health record (EHR) environment, assessing the ability to identify individuals with high-risk. Together, we sought to demonstrate how patient-reported language can be transformed into interpretable and actionable signals that can inform clinicians for earlier stroke risk detection among those with elevated risk.

**Results**

We identified 71,358 individuals with diabetes and 28.5% (n=20,312) experienced strokes in the past decade. Among those diagnosed with diabetes first and then stroke afterwards, we included individuals sent at least one message in our study (22.0%, n=4,465/20,312) along with their secure messages (n=29,084) (Figure 2). We included a matched control of 1,947 with their secure messages (n=13,258).

The majority of patients with stroke were 65 years and older (83.8%, n=3,741/4,463). Nearly half were of a non-White race (47.2%, n=2,106/4,465), and Hispanics were about 13.5% (n=603/4,465). Approximately one-third of the patients were unmarried (37.5%, n=2,788/4,463), and sex was mostly balanced (45.0% female, n=2,010/4,465). Most individuals in the control group were aged 65 and older (99.5%, n=1,938/1,947), and race, ethnicity, sex, and marital status were well matched (Table 1).

*Iterative LLM-Guided Topic Mapping*

We identified 35 main topics, 109 sub1 topics, and 495 sub2 topics from LLM-guided topic mapping of 29,084 messages (Supplementary Table 1). Among those, 65.7% (n=325/495) were clinical concerns, including musculoskeletal (e.g., injuries, mobility, and chronic pain), cardiovascular (e.g., monitoring of blood pressure, heart rate, vitals, and management of atrial fibrillation, anticoagulation, lipids), and neurological (e.g., dizziness/vertigo, cognitive decline, gait/balance issues) concerns across 16 specialty areas. Two investigators' agreement regarding LLM's reasoning for topic clustering was high (Gwet's AC1, 0.93). Topic clusters from BERTopic and LLM-guided taxonomy building were overlapped about 60%, demonstrating noisy nature of patient messages (40% of messages were unassigned) yet powerful semantic reasoning of LLM. Two investigators' agreements for LLM's annotation were high (Gwet's AC1, 0.97 in main topic labeling and 0.88 in symptom specific labeling).

*Risky Symptoms Ranked by Statistical Relevance and Temporal Proximity*

Based on statistical relevance and temporal proximity, we assigned the symptoms to four risk tiers; high, moderate, moderate-low, and low risks (Supplementary Table 2). We categorized high-risk symptoms into four clinical domains: classic vascular risk (e.g., blood pressure management and cardiologic testing inquiries and results), frailty and functional decline (e.g., joint/muscle stiffness/pain/cramps and falls/fractures), early neurological manifestations (e.g., dizziness/vertigo, nausea/vomiting, and migraine), and acute biological triggers (e.g., cold/flu-like symptoms, vaccines/immunization related concerns) (Figure 3). The moderate risk symptoms were categorized into five domains, including aging/neurodegenerative and multi-comorbidities (e.g., dementia/cognitive decline and cysts/moles/skin growth), multi-system disease and end-organ damage (e.g., prostate issues and wound/blisters/skin ulcers), complex diabetes management (e.g., diabetes treatment/insulin pump/continuous glucose monitoring (CGM) device and diet/exercise), musculoskeletal burden (e.g., back pain, ankle/foot/toe pain/swelling, and sciatica), and infection and systematic inflammatory triggers (e.g., chills/fevers and asthma). Figure 4 shows the list of symptoms by risk levels, along with event association score and short-term temporal scores.

*Hybrid Stroke Risk Screening Simulation*

We included 69 patients with diabetes in 3-day (94 messages), 117 in 7-day (n=184messages), 178 in 14-day (n=368 messages), 261 in 30-day (n=795 messages), 372 in 60-day (n=1,449 messages), and 405 in 90-day screening (n=1,860 messages) (Figure 5). We achieved a very high specificity (1.00) and prevalence adjusted PPV (1.00), as intended, across 3-90 day screening windows. Sensitivity was mostly consistent across time blocks while 3-day sensitivity varied likely due to small sample size. Overall sensitivity was highest in 90-day screening (0.63-0.72) followed by 60-day screening (0.59-0.69),whereas shorter screening window showed lower sensitivity (7-day, 0.45-0.50; 14-day, 0.35-0.42). Alert burden ranged 0.16-0.35, where 0.35 was interpreted as 35% of patients will get at least one alert among those evaluated. A full performance metrics with sample size for temporal blocks is in Supplementary Table 3.

**Discussion**

In this study, we developed and evaluated an LLM-guided framework to improve early stroke risk recognition using patient-reported symptoms among individuals with diabetes. By constructing a symptom taxonomy grounded in patients' own language and applying a dual ML pipeline (a heterogeneous GNN and a widely validated EN/LASSO), we identified clinically meaningful symptom patterns associated with subsequent stroke. We then translated these findings into a hybrid risk screening system that integrates symptom relevance and temporal proximity, and evaluated its performance through EHR-based temporal simulation across 3- to 90-day windows with an explicit focus on high specificity. Using this approach, we identified coherent high-risk symptom domains, including classic vascular risk, frailty and functional decline, early neurological manifestations, and acute biological triggers. Under conservative thresholds, deliberately designed to minimize false alerts, the screening system achieved very high specificity and prevalence-adjusted PPV, with moderate sensitivity, an expected trade-off due to the prioritization of precision, that was highest in the 60- to 90-day window. Together, the findings demonstrate that patient-reported language alone can support high-precision, yet, low-burden early stroke risk detection, which could offer a valuable time window for clinical evaluation and close monitoring for immediate intervention when required, holding promising potential in timely stroke care for high-risk individuals.

High-risk symptom clusters were characterized by both significant relevance and temporal proximity, including established early warning signs (e.g., hypertension, cardiac issues), known triggers (e.g., flu-like symptoms), and clinically reasonable but often overlooked indicators (e.g., frailty), falling into four conceptual domains. First, classic vascular risk-related issues, including managing blood pressure, A1C goals and monitoring, anticoagulations and clots, and cardiac testing and surgery. Hypertension is the leading modifiable cause of stroke, accounting for nearly half of all cases.[13–15] Managing blood pressure likely captures uncontrolled blood pressure, consistent with the vascular instability prior to stroke. Simialry, poor blood glucose control is a strong risk factor for stroke in those with type 2 diabetes, showing a dose-response relationship.[16,17] Discussing glycemic control likely grappled with a current suboptimal vascular state. Messaging about anticoagulation, clots, and deep vein thronbosis (DVT) likely marks patients with underlying cardioembolic risks, and a suboptimal international normalized ratio (INR) range is significantly associated with substantially high stroke rates, especially in atrial fibrillation.[18] Discussions of cardiac testing (e.g., ECG/EKG, Holter, Echo) suggest evaluation for subclinical cardiac conditions shortly before stroke, consistent with cardioembolic burden in this high-risk diabetes cohort.[19]

Second topic cluster is acute biological triggers (e.g., cold/flu-like symptoms, coughs, vaccine-related, appetite/taste changes). Common infections, including influenza-like illnesses, increased the odds of stroke days to months after infection, via inflammation, hemodynamic stress (e.g., fever, dehydration), and pro-thrombotic changes, suggesting that the patient-reported respiratory illness could be a meaningful pre-stroke signal.[20,21] Vaccine-related messages probably act as a proxy for high-risk patients, infection outbreaks or histories, and increased healthcare contact.[22] Loss of taste and smell, along with changes in appetite, are key symptoms of viral illnesses like COVID-19, which is associated with increased risk of ischemic stroke.[23]

Third topic cluster is early or atypical neurological manifestations, including dizziness/vertigo, nausea/vomiting, and headaches/migraines. Posterior circulation strokes commonly present with dizziness or vertigo, yet are disproportionately misdiagnosed as benign vestibular conditions.[24] Similarly, nausea and vomiting are classic but often under-appreciated symptoms of posterior circulation stroke, being misattributed to gastrointestinal issues that lead to the delayed diagnosis.[25] Hence, dizziness/vertigo and nausea/vomiting messages are possibly capturing early posterior circulation ischemia or transient ischemic attack (TIA)-like symptoms. Multiple meta-analyses reported that migraine is associated with an increased risk of ischemic stroke,[26] so patient-reported headaches and migraines may have captured a combination of migraine biology, vascular risk, and potential early cerebrovascular events.

Fourth topic cluster is frailty and functional decline, including joint/muscle pain, gait concerns, falls/fractures, and malaise/fatigue. Musculoskeletal pain and stiffness may reflect frailty, reduced mobility, and chronic inflammatory conditions, all related to higher cardiovascular risk. Slow gait speed and gait disorders are closely linked to cerebrovascular disease. Partiuclary, frailty and imparied mobility are increasingly recognized as important modifiers of stroke risk and prognosis in older adults.[27] Falls/fracture messages likely captured pre-stroke gait instability, balance problems, and neurodegenerative changes manifest first as falls before a major cerebrovascular event is recognized.[28] Lastly, self-reported fatigue was associated with up to a 50% increased incidence of stroke.[29] Our graph-augmented LLM pipeline essentially recovered clinically meaningful pre-stroke signals aligned with existing literature, yet from patients' own language.

Moderate risk symptom clusters reflect either significant relevance or temporal proximity, in general within the context of systemic vulnerability, yet they can be mapped onto four conceptual domains, including multisystem vascular and end-organ damage (e.g., renal function, wound/ulcers, prostate, endocrine/metabolic issues), infection and systemic inflammatory triggers[21] (e.g., dysuria/UTI, chills/fever, nasal congestion), frailty and musculoskeletal burden[30] (e.g., back pain, sciatica, neck/shoulder pain, foot/ankle pain, tremors), and complex diabetes management and lifestyle discussions. Chronic kidney disease is an independent risk factor for stroke, and prostate symptoms in the context of lower urinary tract symptoms in males are associated with increased CVD risk.[31] The messages about broader endocrine issues (e.g., thyroid function and dyslipidemia) beyond diabetes and hypertension likely captured complex metabolic conditions, which could be a slow and background driver of stroke. Chronic back pain is closely related to frailty, reduced physical activity, and disability. And in turn, frailty is strongly associated with poor cardiovascular outcomes.[32] Sciatica and radicular leg pain usually reflect reduced mobility, chronic pain, and sometimes falls. Diabetic foot ulcer along with ankle, foot, and toe pain and swelling, these painful lower extremity conditions could be consistent systemic vascular risk markers.[33] Furthermore, we also captured discussions of cognitive decline and disorientation, a known predictor of stroke,[34] as moderate risk symptoms, which could signal the advanced small-vessel and neurodegenerative conditions of the patients.

Our stroke risk screening simulation achieved very high specificity and prevalence adjusted PPV (1.00), meaning every alert corresponded to a true stroke case with no false positives. We intentionally designed our hybrid screening system to be highly conservative, prioritizing certainty over sensitivity. While this trade-off reduced sensitivity, the system could still capture two in three individuals at high risk of stroke with patient-reported signs and symptoms alone, 2-3 months in advance. This conservative approach was essential for clinical deployment to minimize unnecessary alert and prevent increased clinician burden, ensuring that when the system flags a patient, clinicians can trust the alert represents genuine stroke risk requiring immediate attention. Early risk alerts could empower patients by helping them prepare for emergencies and encouraging them to manage known risk factors (e.g., A1c, blood pressure, and cholesterol, ABC) through medications and monitoring devices. Alerts could also assist clinicians to monitor patients' ABC closely. Together, early warning system could help at-risk patients and clinicians be ready for emergencies, potentially overcoming early risk recognition gap and addressing patients' symptom uncertainty, a critical step for timely response in stroke. Future directions include evaluating clinicians' acceptance of alerts (whether to take them seriously or ignore), workflow design in operation (who receives alerts and is responsible), and perceived burden (cognitive and EHR time). Assessing patients' perspectives on when and how they prefer to be informed about their potential stroke risks could guide further workflow design and clinical implementation.

This study has limitations. First, we did not perform external validation. However, we internally validated our risk screening system using three time series patient cohorts from new patient groups, which differed from those used for symptom taxonomy building and the dual ML pipeline. Moreover, our patient cohort came from 22 community practice partners, representing a range of sites, in addition to a large academic hospital and included diverse patient groups; nearly half were non-White races or females. Future external validation of stroke risk screening in other geographic regions is necessary to enhance the generalizability of the approach. Second, selection bias could be possible as our study could only include patient message users. However, individuals with diabetes were among the most active users of secure messaging, and at SHC, more than 70% of patients with diabetes already used it a decade ago.[7] Thus, selection bias could be minimal. The added value of our study includes that while most stroke-prediction studies have focused on structured EHR data, vitals, lab results, and imaging, our study demonstrated that patient-reported signs and symptoms from secure messaging could be a valid source to detect the risk of stroke, adding a new data venue to the field. Currently, many health systems already have millions of patient messages, yet those were rarely used for predictions. We demonstrated the feasibility of passive stroke screening for early symptom detection with high precision in a real-world digital health setting (patient portal). Our study showed that the existing infrastructure (patient portal) could serve as a passive warning system for at-risk individuals for stroke, and our framework could be further applied to other early symptom detection. The findings could open novel opportunities for automated triage and proactive patient outreach, and clinical decision support. Second, while previous studies established risk factors of stroke, our work quantified the temporal proximity of those symptoms from days to weeks before stroke, adding temporal granularity to our understanding of how patients' body deteriorates before acute cerebrovascular injury illustrated by patient themselves. Third, our dual ML framework for identifying clinically meaningful symptom categories in patient messages is methodologically robust and novel. It accounted for the complex relationships among symptoms, patients, and events and provided a clinically interpretable risk score, overcoming a common limitation of ML: interpretability. Fourth, our risky symptom categories captured known prodromal conditions, known triggers, and under-recognized yet clinically plausible predictors. This could broaden understanding of stroke symptom presentation to a multidimensional lens from classic warning signs. Lastly, our hierarchical symptom taxonomy, explicitly derived from patients' own expressions, provides a comprehensive overview of patients' concerns. With this patient-centered symptom ontology, it could enable standardized symptom-based surveillance across health systems. The LLM-guided taxonomy-building pipeline can be readily reproduced and applied to other diseases.

Stroke is a significant health problem worldwide, and early risk detection is critical. The patient-centered, AI-guided passive surveillance system demonstrated the ability to capture at-risk individuals with high certainty a few months earlier, which could offer a valuable time window for clinical evaluation and intervention in high-risk individuals and holding potential for timely stroke care moving from reactive care to proactive care.

## Methods

Patient-Centered, LLM-Guided Early Stroke Risk Screening Framework

Figure 1 demonstrates envisioned care flow of integrating the passive surveillance for early stroke risk to the healthcare system, moving from reactive care (where care starts when stroke incidence happens) to proactive care (where care starts before stroke incidence by empowering patients and assisting clinicians).

Data Source and Patient Cohort Identification

We identified and included patients diagnosed with diabetes (ICD-10 codes: E08, E09, E10, E11, E13) temporally prior to diagnosis with ischemic stroke (ICD-10: I63) in Stanford Health Center (SHC) in the past 10 years (08/2014-07/2024). We identified those with diabetes but no associated stroke diagnosis through exact stratified frequency matching for sociodemographic factors (age, race, ethnicity, sex, and marital status) as a matched control.[35] For symptom discovery via dual ML pipeline, we oversampled case cohorts to ensure sufficient stroke features for stable estimation of symptom association as stroke is rare event in real world. For screening simulations, we sampled case and control as 1:1. We obtained de-identified secure patient messages (Patient Medical Advice Requests) by Safe Habor method,[36] routed to primary care, internal medicine, and family medicine at SHC and 22 community practice partners in Northern California. We conducted all analyses in a HIPAA-compliant analytics environment. The Stanford University Institutional Review Board approved this study.

Iterative LLM-Guided Topic Mapping

To characterize patients' clinical concerns, we developed an iterative LLM-guided taxonomy-building pipeline. We bootstrapped the initial 10 main categories from a seed set of messages (n=200) using LLM (Gemini Pro 3, November 18, 2025, Google DeepMind LLC), then streamed the remaining corpus in small batches (n=50 messages) to create a three-level hierarchy (MAIN→SUB1→SUB2) that balances clinical interpretability and topical granularity. We instructed the LLM to preserve existing topics whenever possible, adding new topics only when incoming messages present clearly novel issues and merging existing topics only when nearly identical. When the LLM updated the topics, it recorded newly added or merged topics with brief reasons. Two investigators independently reviewed these changes and reported Gwet's AC1, due to expected imbalanced class (e.g., agree > disagree), to show agreement for validation.[37] We additionally validated the LLM generated topic clusters using an established topic modeling (BERTopic) with a small efficient embedding (all-MiniLM-L6).[38] We compared the overlap of main topics from LLM-guided topic mapping and BERTopic. Full detailed methods and prompts for LLM-guided topic mapping, and validation are available in Supplementary Methods 1 and 2, respectively. We then, using symptom taxonomy, converted text message data to a structured data using LLM (Gemini3, Google DeepMind, LLC) that can be used for ML pipeline in the next step. Two investigators rated LLM's annotation for small subset (n=200 messages, 0.5% of total messages from those with and without stroke), and reported Gwet's AC1 for agreement. Full details and prompts used for LLM annotation are in Supplementary Methods 3.

Graph Augmented LLM
To identify clinically meaningful symptoms associated with stroke, we developed a dual ML pipeline leveraging Elastic Net (EN)/Least Absolute Shrinkage and Selection Operator (LASSO) logistic regression, a solid approach to identify predictor symptoms, and a Graph Neural Network (GNN) with LLM support to capture heterogeneous tree-like relationships among symptoms, which EN/LASSO might miss. Our pipeline included patient demographics, comorbidities, and timestamped secure messages annotated by LLM for SUB2 topics. We applied embedding-based semantic similarity (Sentence Transformer) to the LLM-labelled symptoms as an additional edge builder to enhance robustness of GNN.[39] We trained a heterogeneous GNN (GraphSAGE-based HeteroConv layers) to optimize patient-level event prediction by computing a weighted sum of event losses via Adam optimization.[40] We constructed three dual ML (GNN and EN/LASSO) pipelines, symptom-only, symptom with demographics and comorbidities, and symptoms with GNN-predicted risk. In the third pipeline, we updated graphs with temporal (symptom timestamp) and relational (explicit comorbidity nodes and patient-patient similarity nodes) structures, increasing minimum relevance score for message pruning to reduce noise and dropout rates to avoid graph overfitting. We evaluated the discriminative performance (ROC AUC) of three pipelines via 5-fold cross-validation to find the best model.

To estimate symptom importance, we applied two complementary approaches: graph-based salience (prevalence, temporal dynamics, semantic intensity, and centrality of symptoms) and event-based attribution (GNN event delta, changes in patient-level event loss by removing each symptom, and EN permutation score). We computed patient-level symptom importance, representing statistical relevance to stroke, by z-score averaging GNN event delta and EN permutation score (event association score). Full code for a dual ML pipeline is available in Supplementary Method 4.

Hybrid Stroke Risk Screening Simulation with High Specificity Optimization

To evaluate clinical impact of LLM-guided stroke risk screening, we performed a stroke risk screening simulation in the EHR on individuals with diabetes (08/2024-07/2025, SHC), using 3-day, 7-day, 14-day, 30-day, 60-day, and 90-day windows. To ensure robustness and evaluate internal consistency of the performance, we performed temporal validation using three time blocks (08/2024-11/2024, 12/2025-03/2025, 04/2025-07/2025). We included individuals who sent at least one message within each screening window prior to stroke. For control, we randomly sampled the same number of demographically matched individuals with case in each temporal block and followed it throughout each screening window.

We designed a risk screening pipeline by combining two independent signals: how strongly a symptom is associated with future stroke (event association score), and how often that symptom occurs close in time to a stroke event in real patient histories (short-term temporal score, a weighted sum of the probabilities of symptom appearing within 7 days [0.66 weighted] and 30 days [0.33 weighted] of stroke). We categorized symptoms based on the composite screening score, a weighted sum of event association score (0.6 weighted) and short-term temporal score (0.4 weighted), into four risk levels: high, moderate, moderate-low, and low, adding a very-high level for substantial short-term signal (top 15% of short-term temporal score).

We intentionally prioritized high specificity (>0.9) to minimize false-positive alerts in optimizing this hybrid screening model. We implemented a hybrid screening algorithm by combining an interpretable symptom-count rule with a logistic-regression risk score. A patient was flagged as screen-positive if they met symptom-based rule during the screening window or if the probability of stroke exceeded the optimal threshold, identified through our threshold scan (probability range from 0 to 1 in 0.01 increments). We applied an LLM (GPT-5, OpenAI LLC) to screen patient messages for risk symptoms and computed sensitivity, specificity, prevalence adjusted positive predictive value (PPV) and negative predictive value (NPV), assuming deployment prevalence as 0.10, F-1 score, and alert burden (patient-level fraction flagged=number of patients with screening positive/total number of patients evaluated) for each screening window. The full code for the hybrid screening model is provided in Supplementary Method 5.

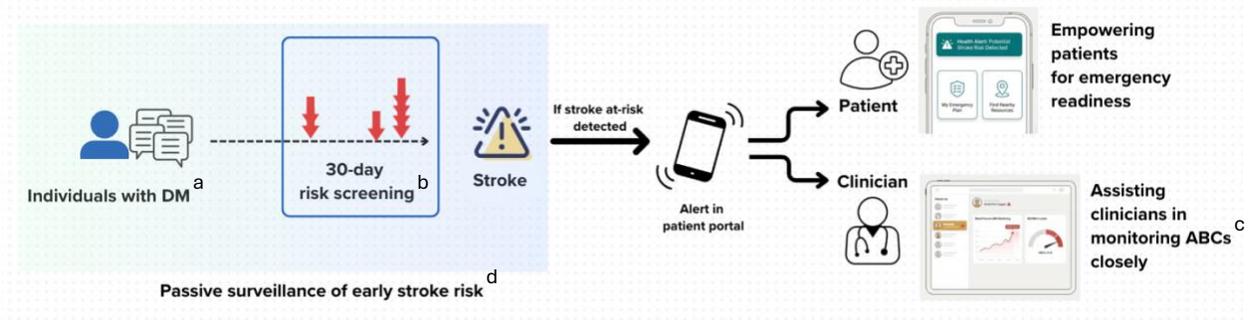

**Figure 1. Conceptual Framework for Patient-Centered LLM-Guided Passive Surveillance System for Early Stroke Risk Detection**

a. DM: Diabetes; b. Screening window could be tailored to each institution; c. ABC: A1C, Blood pressure, and Cholesterol; d. Current study evaluated the passive surveillance system.

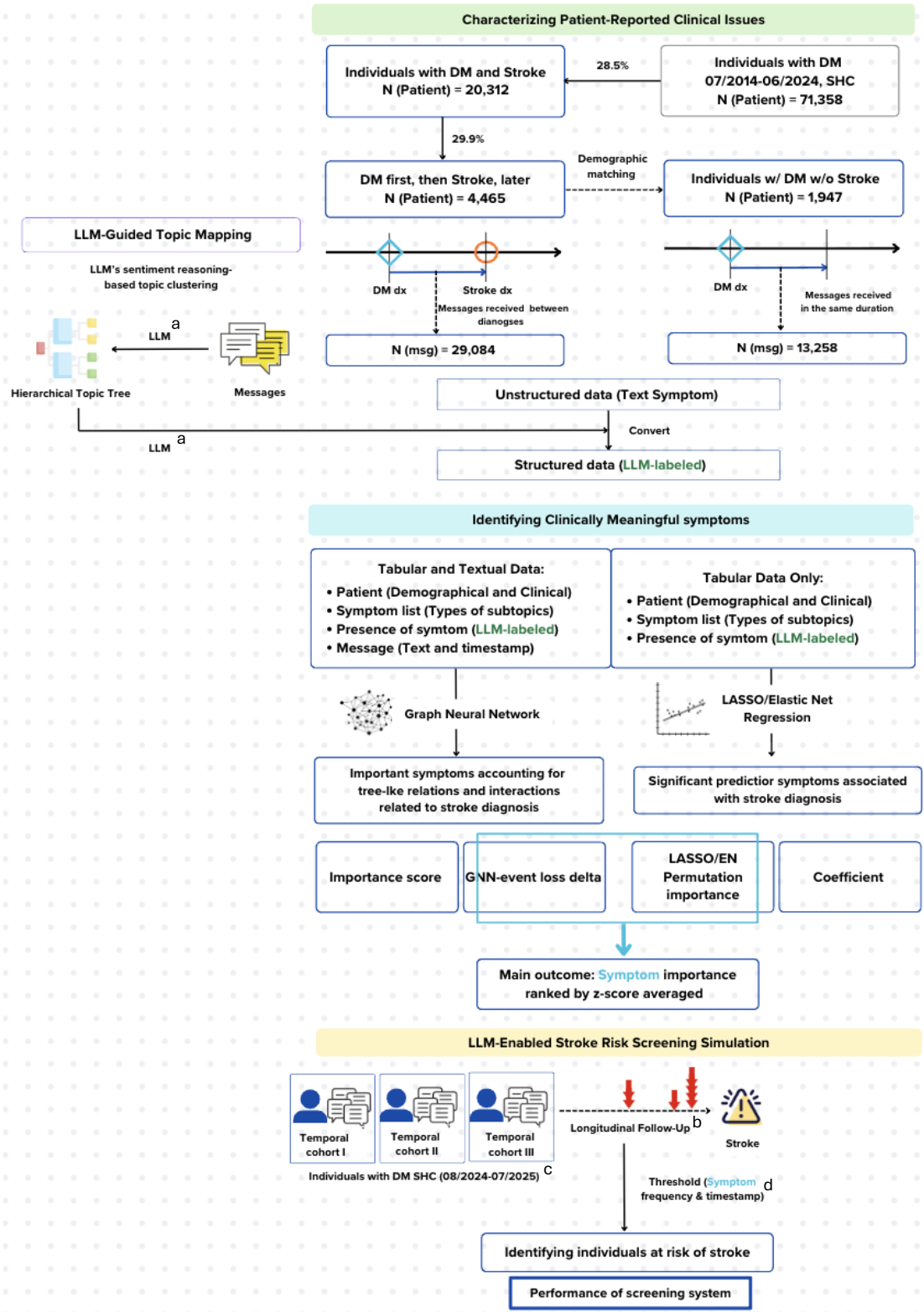

**Figure 2. Study Design and Data Flow**

a. LLM: Gemini3 (Google DeepMind, LLC); b. 3-day, 7-day, 14-day, 30-day, 60-day, 90-day screening window was applied; c. Cohort I (08/2024-11/2024), Cohort II (12/2025-03/2025), Cohort III (04/2025-07/2025); d. Threshold: We provided threshold for optimal outcome (=high specificity) through grid search

**Table 1. Sociodemographic characteristics of patients with diabetes (SHC, 2014-2024)**

|  | Patients with diabetes and stroke[a] (N=4,465) |  | Patients with diabetes without stroke[b] (N=1,947) |  |
|---|---|---|---|---|
|  | Frequency (N) | (%) | Frequency (N) | (%) |
| **Age** | | | | |
|   18-34 | 12 | 0.3 | 9 | 0.5 |
|   35-49 | 114 | 2.6 | 0 | 0.0 |
|   50-64 | 598 | 13.4 | 0 | 0.0 |
|   65-74 | 1094 | 24.5 | 577 | 29.6 |
|   75+ | 2647 | 59.3 | 1361 | 69.9 |
| **Race** | | | | |
|   Asian/Pacific Islander | 1010 | 22.6 | 446 | 22.9 |
|   Black | 326 | 5.3 | 129 | 6.6 |
|   White | 2271 | 50.9 | 1010 | 51.9 |
|   Native American/Other | 770 | 17.2 | 316 | 16.2 |
|   Unknown | 88 | 2.0 | 46 | 2.4 |
| **Ethnicity** | | | | |
|   Hispanic/Latino | 603 | 13.5 | 237 | 12.2 |
|   Non-Hispanic | 3727 | 83.5 | 1637 | 84.1 |
|   Unknown | 135 | 3.0 | 73 | 3.7 |
| **Sex** | | | | |
|   Female | 2010 | 45.0 | 868 | 44.6 |
|   Male | 2454 | 55.0 | 1078 | 55.4 |
|   Unknown | 1 | 0.0 | 1 | 0.1 |
| **Marital Status** | | | | |
|   Married | 2788 | 62.5 | 1206 | 61.9 |
|   Unmarried[c] | 1675 | 37.5 | 726 | 37.3 |
|   Unknown | 27 | 0.6 | 15 | 0.8 |

a. Individuals were identified by ICD-10 codes for diabetes and stroke (Stanford Health Care, 2014-2024); b. Demographic matching: We used simple random sampling without replacement from the demographic-matched control group; c. Unmarried included divorced, separated, widowed, or single/never been married, or other

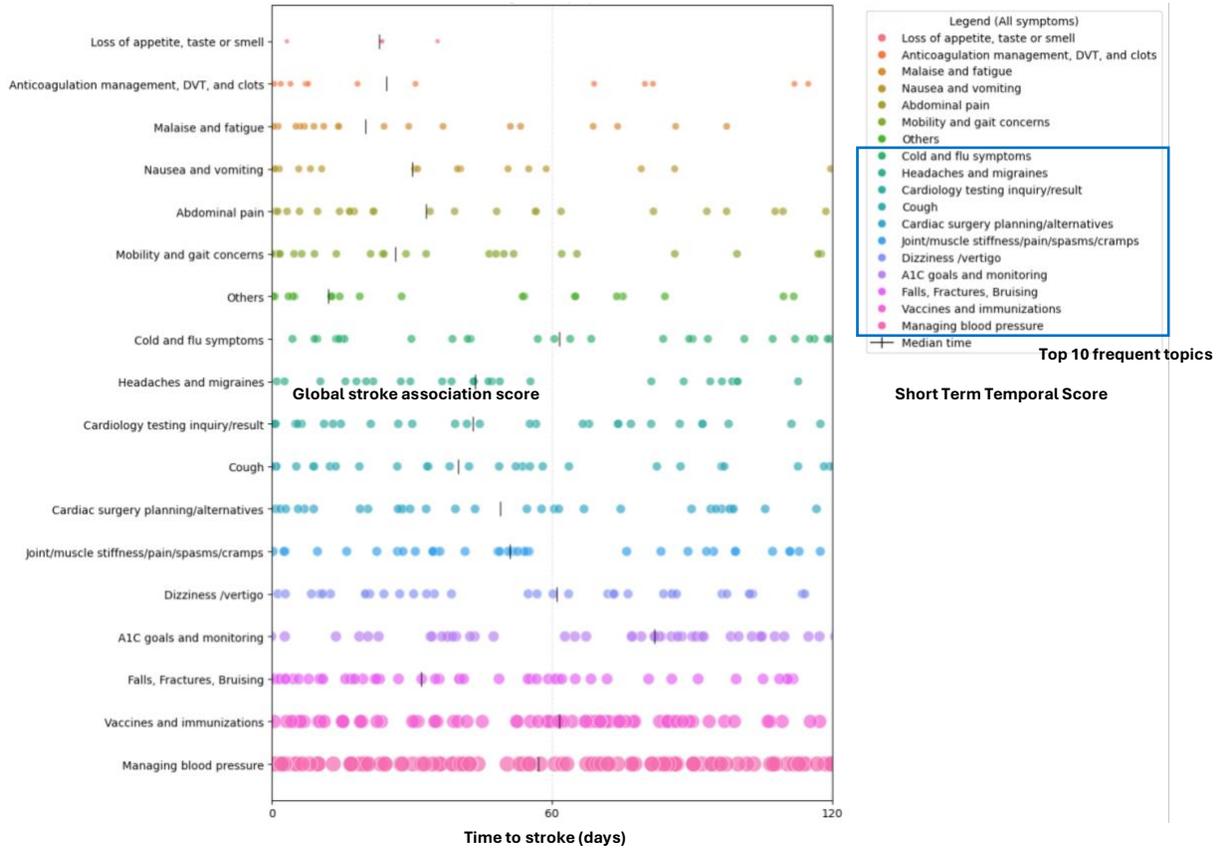

a. Frequency of high-risk symptoms by time to stroke

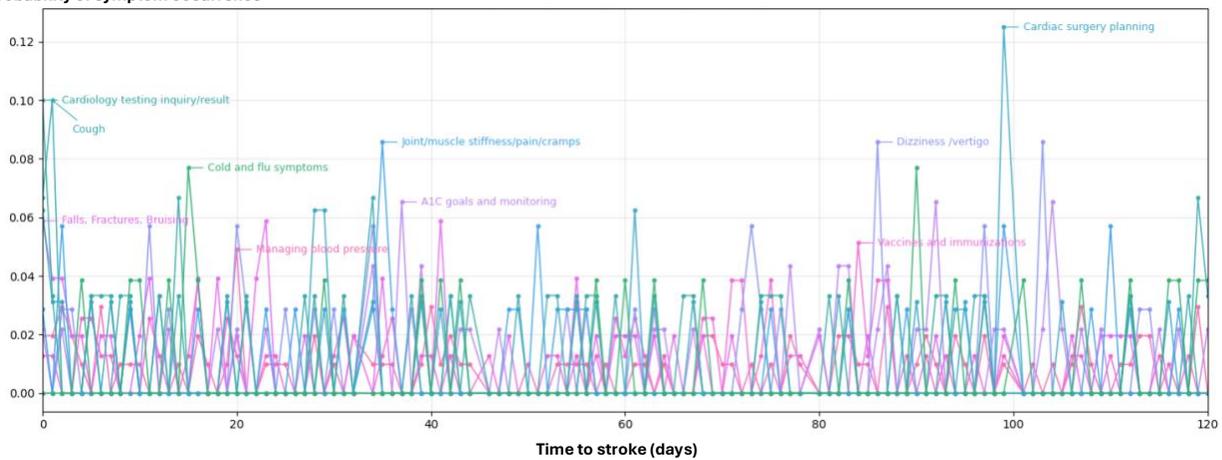

b. Proximity curve of high-risk symptoms by time to stroke

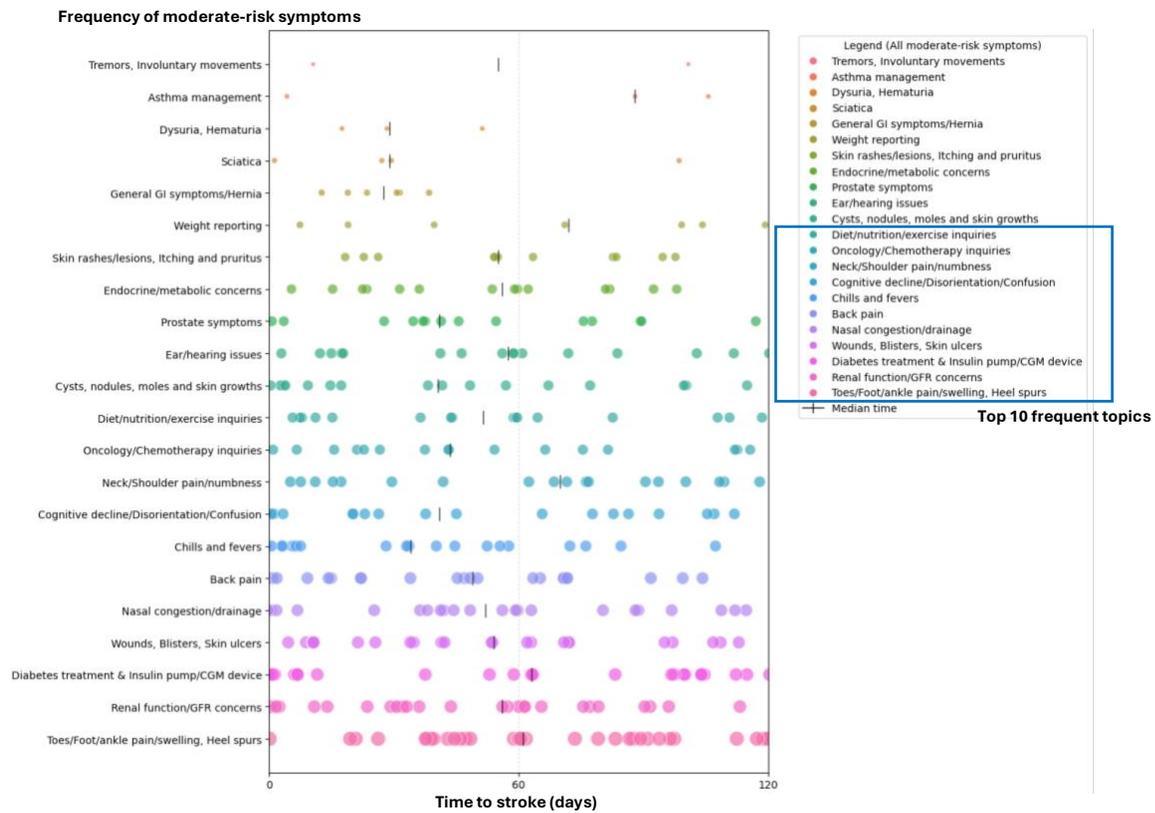

c. Frequency of moderate-risk symptoms by time to stroke

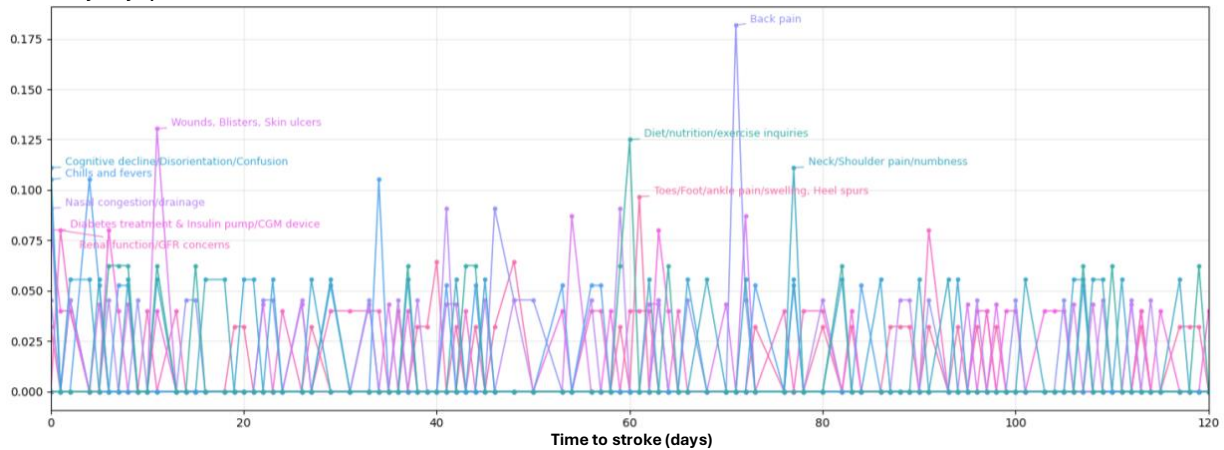

d. Proximity curve of moderate-risk symptoms by time to stroke

**Figure 3. Temporal Proximity of Symptoms with High or Moderate Risks**
We plotted high-risk symptom frequency (a) and moderate-risk symptom frequency (c) by time-to-stroke until 120 days of those with diabetes and subsequent stroke at SHC 2014-2024 to show frequency distribution by time; We plotted probability of symptom occurrence (probability of symptom A occurrence =number of symptom A messages in a given day/total number of symptom A messages in the observational window) for top 10 frequent high-risk symptoms (b) and top 10 frequent moderate-risk symptoms (d) by time to stroke of those with diabetes and subsequent stroke to show probability distribution of symptom appearance by time.

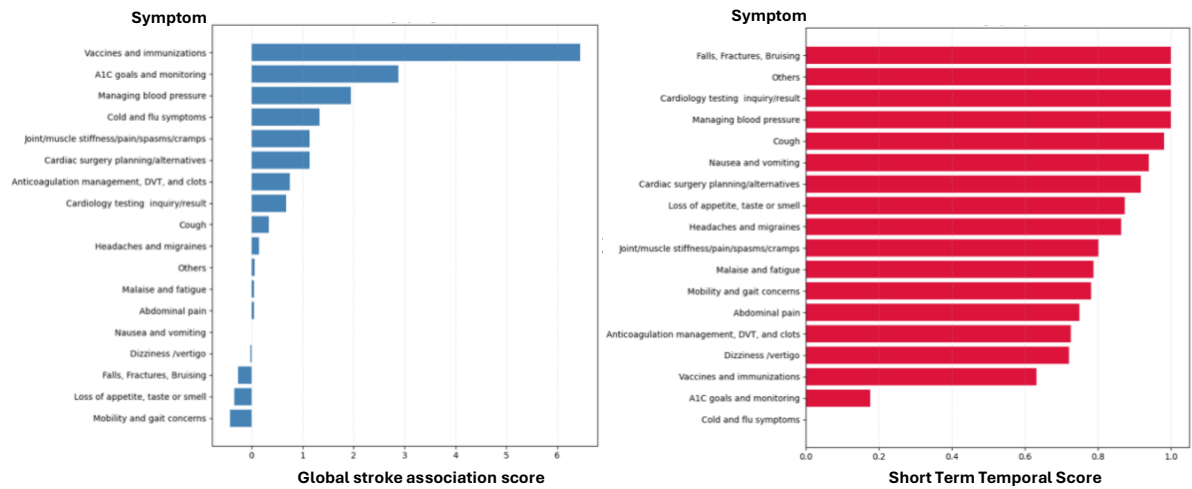

a. Global stroke association score by high-risk symptom

b. Short Term Temporal Score by high-risk symptom

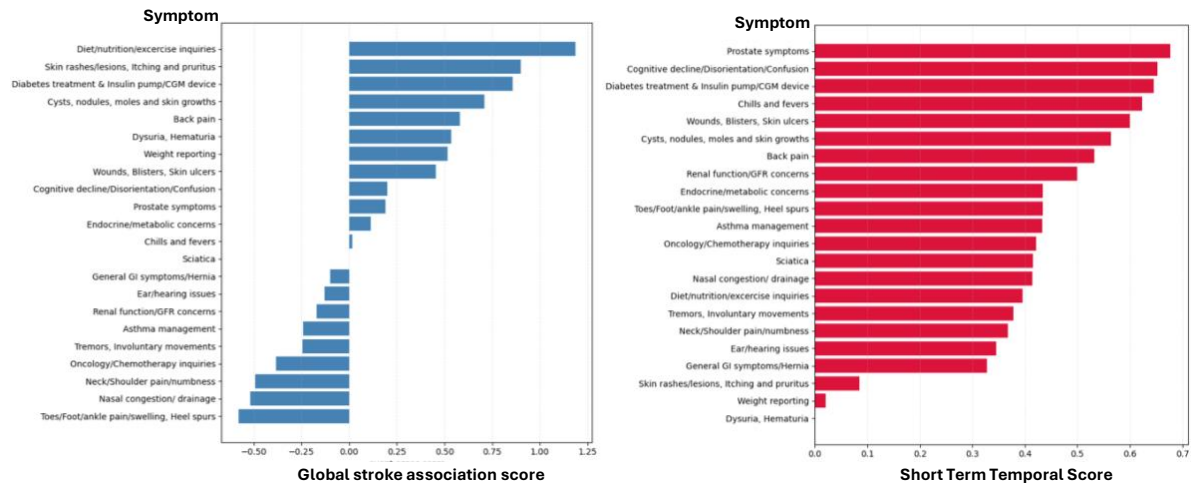

c. Global stroke association score by moderate-risk symptom

d. Short Term Temporal Score by moderate-risk symptom

**Figure 4. Symptoms with High or Moderate Risks Ranked by Statistical Relevance (Global Stroke Association Score) and Temporal Proximity (Short Term Temporal Score)**

a. Global Stroke Association Score: Z-score averaged of graph neural network event loss delta and least absolute shrinkage and selection operator/elastic net permutation score, representing a significant association with and importance to stroke; b. Short Term Temporal Score: A weighted sum of pct 7 and pct 30 that can represent short-term temporal proximity to stroke (=0.66 x pct 7 + 0.33 x pct 30) where pct 7 was the probability of the symptom occurring within 7 days prior to stroke and pct 30 was the probability of the symptom occurring within 30 days prior to stroke. Full symptom list by risk category (high, moderate, low) with global stroke association score and short term temporal score is available in Supplementary Table 2.

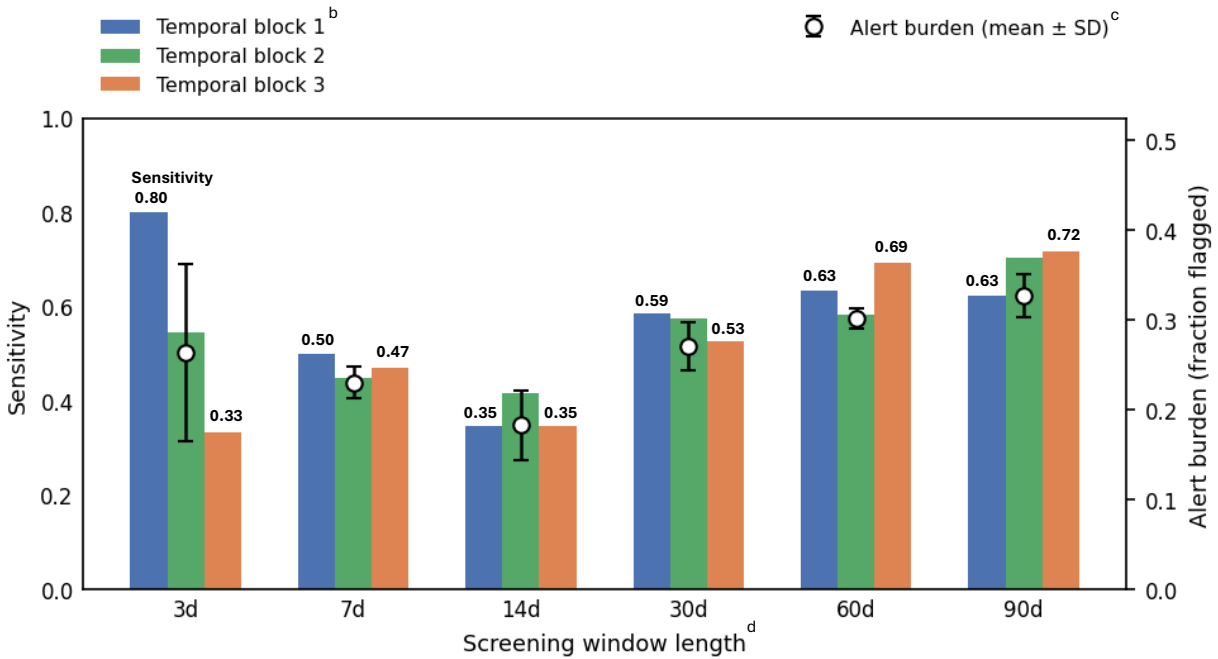

**Figure 5. Sensitivity Across Temporal Validation Blocks with Alert Burden of Stroke Risk Screening system[a]**

a. We intentionally designed a hybrid risk screening system to achieve high specificity (>0.90) via grid search optimization rather than maximizing the F-1 score. Hence, shown sensitivity was obtained under high-specificity operating threshold where we obtained very high specificity of 1.00 (=how many are correctly classified as low risk for those who will not have a stroke) and prevalence adjusted PPV of 1.00 (=how many are actually having a stroke among those flagged as high risk); b. Three grouped bar graphs represent sensitivity (y-axis, left) for each temporal block (Temporal block 1: 08-01-2024 to 11-31-2024; Temporal block 2: 12-01-2024 to 03-31-2025; Temporal block 3: 04-01-2025 to 07-31-2025) across screening windows (3-day to 90-day). Sensitivity was mostly consistent for three-time blocks in each screening window, except for 3-day screening (widely varied perhaps due to small sample size), and increased with longer screening windows; c. White dots represent mean alert burden (y-axis, right) with standard deviation (vertical line) for three temporal blocks across screening window. Mean alert burden also increased with longer careening window (highest in 90-day screening, mean=0.328, SD=0.024; lowest in 14-day screening, mean=0.183, SD=0.038); d. sample size for 3-day (n=33 patients with stroke [n=42 messages], n=36 patient without stroke [n=52 messages]), 7-day (n=57 patients with stroke [n=85 messages], n=60 patient without stroke [n=99 messages]), 14-day (n=88 patients with stroke [n=193 messages], n=90 patient without stroke [n=175 messages]), 30-day (n=126 patients with stroke [n=443 messages], n=135 patient without stroke [n=352 messages]), 60-day (n=177 patients with stroke [n=777 messages], n=195 patient without stroke [n=672 messages]), and 90-day (n=196 patients with stroke [n=987 messages], n=209 patient without stroke [n=873 messages]). A full performance metrics with sample size for temporal blocks is in Supplementary Table 3.